%% file: spark_main.tex
\documentclass{article} % For LaTeX2e
\usepackage{iclr2027_conference,times}

\input{math_commands.tex}

\usepackage{graphicx}
\usepackage{xcolor}
\usepackage{hyperref}
\hypersetup{colorlinks=true,linkcolor=black,citecolor=black,urlcolor=black}
\usepackage{url}
\usepackage{enumitem}
\usepackage{booktabs}
\usepackage{multirow}
\usepackage{algorithm}
\usepackage{algpseudocode}
\usepackage{tikz}
\usetikzlibrary{arrows.meta,positioning,fit,backgrounds,calc}

\title{SPARK: Self-Correcting Embodied Memory\\via Physical Re-Observation}

\author{Dmytro Kurdydyk$^{1,\dagger}$ \quad Aadi Palnitkar$^{2}$ \quad Sebastian Scherer$^{3}$ \quad Seungchan Kim$^{3,\dagger}$ \\
$^{1}$Davidson College \quad $^{2}$University of Maryland \quad $^{3}$Carnegie Mellon University \\
{\small \texttt{dmkurdydyk@davidson.edu, seungch2@andrew.cmu.edu}}}

\iclrfinalcopy

\begin{document}
\maketitle
{\renewcommand{\thefootnote}{\fnsymbol{footnote}}\footnotetext[2]{Corresponding authors.}}
\lhead{Preprint}

\begin{abstract}
Retrieval-augmented generation (RAG) systems have seen growing adoption
in embodied settings as language-queryable memories of the environment.
However, existing approaches often assume a reliable single-episode
knowledge base (KB) or accumulate observations across episodes without
correcting prior errors. Such errors can persist through incorrect
semantic labels, inaccurate object positions, and inconsistent spatial
relationships. To address this, we present SPARK, a corrective
multi-episode framework for embodied RAG. After the first episode, a
semantic critic flags regions of the KB, a targeted planner generates
revisits, and a selective merge adds, removes, or replaces KB information
based on the new observations. We evaluate SPARK on four outdoor AirSim
maps under two degraded viewpoint conditions, using perception-built KBs,
a visibility-constrained oracle used only for evaluation, and a fixed set
of 158 embodied question answering (EQA) questions scored before and
after correction.
A single corrective episode improves downstream EQA on every map
under both conditions while storing 13--41\% fewer objects than
append-only updating, with gains up to $2.7\times$ on one map-condition
pair. Extending correction through Episode~4 brings every map above
the EQA performance of its undegraded reference.
Ablations show that the critic's three signals each contribute to
performance, that targeted revisits achieve a larger mean gain than
untargeted ones under the same budget, and that selective merging
achieves gains comparable to append-only updating with a smaller KB.
Overall, SPARK demonstrates the value of correcting embodied memory
through targeted re-observation, particularly as embodied systems
scale and the risk of accumulated errors increases.
\end{abstract}

\section{Introduction}

Retrieval-augmented generation (RAG) originated in natural language
processing research, where external knowledge bases (KBs) store
information that can be semantically retrieved to augment a
language model's
context \citep{lewis2020rag}. This provides access to non-parametric
knowledge beyond the model's weights and has inspired extensions such
as graph-based RAG \citep{graphrag2024}. More recently, RAG has
extended into robotics, where embodied memory systems organize
multimodal observations into persistent representations, such as vector databases or hierarchical semantic structures, to support
retrieval, reasoning, and long-horizon embodied question answering
(EQA) \citep{embodiedrag2024,remembr2024}.

Embodied memory, however, is built from inherently noisy and
incomplete physical observations \citep{gu2024conceptgraphs}. Perception can introduce spurious or duplicated objects, wrong labels, and inaccurate positions \citep{dengler2021online,matez2022ltc}, which persist in the KB and affect downstream tasks. As embodied
systems scale, a central question emerges:
\emph{how can a robot detect and correct errors in a memory that
has already been built?} In large language model (LLM) research, unreliable retrieved
knowledge has motivated corrective mechanisms such as CRAG and
Self-RAG \citep{crag2024,selfrag2024}. These approaches, however,
focus on retrieval and generation rather than physical
re-observation as a source of corrective evidence.

Object-level semantic mapping maintains existence and class beliefs from later detections \citep{dengler2021online,matez2022ltc}, but language-queryable embodied memories rarely correct an already-built KB, often treating stored knowledge as reliable or incorporating new observations without revising prior errors \citep{embodiedrag2024,remembr2024}.
One line of work reuses what was already captured, through dense
representations such as point clouds and 3D Gaussians or through stored keyframes, to
reinterpret stored observations, retrieve previously captured
frames, or synthesize views from reconstructed scenes
\citep{rag3dsg2026,ogscene3d2026,graphpad2025}.
These methods refine a KB by inference from existing observations
and cannot add physical evidence about regions that were never
adequately observed. Other approaches act physically in the environment, through exploration or corrective action, but serve different purposes. GSMem interleaves physical exploration with answering, exploring further whenever its map cannot answer the current query \citep{gsmem2026}, and Mind Palace
stores each episode as a separate scene-graph world instance and treats differences as environmental change \citep{mindpalace2025}, while
REFLECT summarizes the robot's own execution history to explain and recover from immediate task failures \citep{reflect2023}. What remains missing is a system
that can diagnose an existing KB independently of any particular
query and resolve identified errors through targeted new
observations across episodes.

To address this gap, we present SPARK,
\emph{Self-Corrective Process for Active Robotic Knowledge},
a corrective multi-episode framework for embodied RAG.
SPARK employs a VLM-powered \emph{critic} that flags KB regions
independently of any EQA query, followed by a
targeted planner that generates revisit viewpoints for acquiring
new evidence. A selective merge then adds, removes,
or replaces KB information based on these observations,
closing the loop between diagnosis, re-observation, and
correction across episodes.

We implement SPARK in AirSim \citep{airsim2017} using an
RGB-D perception pipeline combining class-agnostic segmentation,
depth backprojection, and open-vocabulary VLM labeling. We evaluate
downstream EQA before and after correction and separately assess
the critic against a visibility-constrained oracle graph,
constructed using field-of-view restrictions and raycasting
from the robot's observation waypoints. This oracle is used
only for evaluation. We validate SPARK across four outdoor
scenes, two degraded viewpoint conditions, successive
corrective episodes, and ablations of the critic, revisit
policy, and merge policy.

Our \textbf{contributions} are:

\begin{itemize}
    \item SPARK, to our knowledge the first framework where a robot repairs its
    own knowledge base without ground truth, using a semantic critic to
    diagnose memory, a planner to target revisits, and a selective merge
    that removes and supersedes stored content. Correction improves EQA on
    every map and viewpoint condition, and stores $13$--$41\%$ fewer objects than
    appending the same observations.
    \item An evaluation setting for memory correction, pairing a perception
    pipeline with a visibility-constrained oracle used only for evaluation,
    and a fixed EQA set of 158 questions across four
    outdoor AirSim maps. Critic-guided revisits yield a larger mean gain than an untargeted baseline
    under a matched budget, and by the third episode all four maps exceed their
    undegraded reference.
\end{itemize}

\section{Related Work}

\paragraph{Building Embodied Memory.}
Robots must often use past observations for later tasks, relying on dense metric maps that preserve observations as persistent spatial representations \citep{henriques2018mapnet}. As robots increasingly act on natural-language instructions \citep{ahn2022can, driess2023palm, huang2022inner}, memory has shifted toward representations that are semantic and language-grounded as well as spatial. ReMEmbR~\citep{remembr2024} stores captioned observation segments in a flat vector database, while Embodied-RAG~\citep{embodiedrag2024}, which we build on, organizes captioned observations by spatial-semantic clustering into a hierarchy of summarized clusters that supports graph-based retrieval. ConceptGraphs instead builds an open-vocabulary 3D scene graph directly from posed RGB-D frames~\citep{gu2024conceptgraphs}. Across these approaches, memory quality depends on what was encoded during observation, and little attention has been paid to correcting information that was stored incorrectly.

\paragraph{Corrections in Memory Systems.}
In LLM research, correction has been applied to RAG at different stages. CRAG, for instance, considers that retrieved documents may be inaccurate and uses web search as a secondary source of corrective evidence \citep{crag2024}, while Self-RAG learns when to retrieve and to critique its own output against retrieved evidence \citep{selfrag2024}. In robotics, Mind Palace is effectively a graph-based RAG maintained across multiple deployments, turning observations into a sequence of scene-graph world instances and answering queries under the assumption that the environment changes \citep{mindpalace2025}, so discrepancies between instances are treated as changes rather than errors. REFLECT uses a hierarchical summary to diagnose task failure and plan a corrective action \citep{reflect2023}, with correction applied to the task instead of persistent memory. SPARK instead studies how an already-built embodied memory carrying imperfect information can be corrected to improve downstream EQA.

\paragraph{Editable Scene Representations.}
A related line of work edits existing scene representations using previously collected evidence. RAG-3DSG corrects low-confidence captions using synthetic views from the reconstructed point cloud and neighboring high-confidence nodes \citep{rag3dsg2026}, while OGScene3D reassigns low-confidence clusters from nearby high-confidence Gaussians \citep{ogscene3d2026}. Both therefore infer corrections from information already encoded, without acquiring new evidence for under-observed regions. Separately, GraphPad retrieves additional keyframes from an existing scan when the current scene graph cannot answer a query \citep{graphpad2025}, limiting its updates to previously captured information and making them query-driven. GSMem shares our premise that a region may be under-observed, but triggers physical exploration only when a query cannot be answered from a rendered view of its Gaussian-splatting map \citep{gsmem2026}, making exploration a query-time fallback rather than a standing diagnosis of the KB. SPARK diagnoses regions independently of downstream queries and physically revisits those it flags, allowing it to recover information that the original observations never captured adequately.

\section{Method}
\label{sec:method}

\subsection{Problem Formulation}
\label{sec:problem-formulation}

Current persistent memory systems generally assume a single collection pass suffices for downstream EQA, yet a KB built from a perception pipeline can accumulate errors, such as missing objects, incorrect labels, and misplaced geometry, that propagate to downstream tasks unless corrected. We study KB correction over repeated episodes in a static world, where cross-episode mismatches reflect memory errors as opposed to scene changes.

An episode is a single traversal of a set of $N_e$ waypoints, with one observation recorded per waypoint, so coverage is determined by the waypoints without continuous recording.

During episode $e$, the robot records
$
\mathcal{O}^{(e)} = \{\, o_{e,k} \,\}_{k=1}^{N_e},$ where $
o_{e,k} = \big(I_{e,k}, D_{e,k}, \mathbf{p}_{e,k}\big)$: $I$ is the RGB image, $D$ the depth map, and $\mathbf{p}$ the robot pose.
Pose is taken as ground truth because localization is not the subject of this
work. The first episode builds the memory
\begin{equation}
H^{(1)} = f_{\mathrm{build}}\big(\mathcal{O}^{(1)}\big),
\end{equation}
by pooling and clustering observations, so visit order does not affect the result.

\begin{figure*}[t]
\centering
\includegraphics[width=\linewidth]{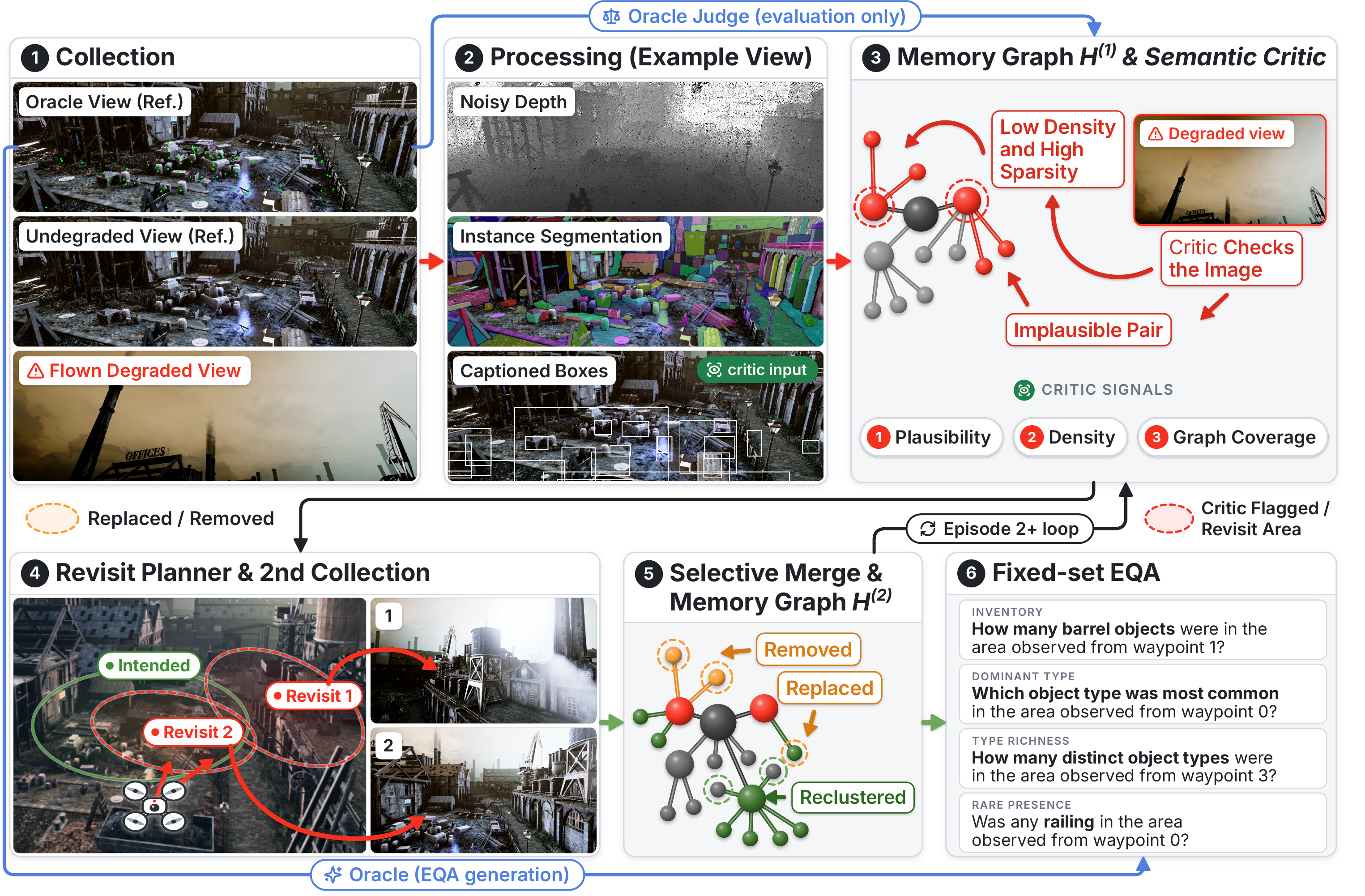}
\caption{\textbf{Overview of the SPARK correction loop.} Episode~1 flies the source
waypoints under a degraded view and builds memory $H^{(1)}$ from its own RGB-D
observations (1--2). The semantic
critic reads $H^{(1)}$ and the Episode~1 images alone, flagging regions by plausibility, density, and graph coverage (3). The planner turns those into viewpoints offset from the source waypoints for Episode~2 (4), and the selective merge removes, replaces, and adds entries to produce $H^{(2)}$ (5), which a
fixed EQA set scores against $H^{(1)}$ (6). The oracle (blue) is evaluation-only and never reaches the critic, planner, or merge.}
\label{fig:loop}
\end{figure*}

Given $H^{(e)}$, a semantic critic $\mathcal{C}$ inspects it together with the
per-episode evidence $Z^{(e)}$ retained alongside it, namely the waypoint images and their provenance, and returns regions flagged as unreliable,
\begin{equation}
\mathcal{R}^{(e)} = \mathcal{C}\big(H^{(e)}, Z^{(e)}\big).
\end{equation}
A revisit planner converts these regions into waypoints,
$\mathcal{V}^{(e+1)} = f_{\mathrm{plan}}(\mathcal{R}^{(e)}, H^{(e)})$, and visiting them yields targeted observations $\mathcal{O}^{(e+1)}$. The memory is then corrected as
\begin{equation}
H^{(e+1)} = f_{\mathrm{correct}}\big(H^{(e)}, \mathcal{O}^{(e+1)}\big).
\end{equation}
The formulation assumes no particular memory representation and requires only a persistent KB, a critic, a planner, and a merge that can remove, replace, and add information (Figure~\ref{fig:loop}). The objective is for $H^{(e+1)}$ to answer questions about the environment better
than $H^{(e)}$. We evaluate the first corrective transition, $H^{(1)}$ to $H^{(2)}$, and study repeated correction separately in
Sec.~\ref{sec:longitudinal}. Ground truth comes from an evaluation-only oracle used for scoring and never available to the deployed system.

\subsection{Perception-Based Memory Construction}

We keep the hierarchical memory representation introduced by
Embodied-RAG~\citep{embodiedrag2024} and illustrated in
Figure~\ref{fig:memory}, since its hierarchy allows corrective updates to act
directly on regions flagged by the critic.
Its released AirSim collection code, however, populates memory from simulator ground truth (AirSim's scene object list or its ground-truth detection API), so every stored object exists and is
correctly labeled and positioned. Our setting requires a KB that can actually be wrong, so we instead build memory from the robot's own sensors: the pipeline receives only synchronized RGB $I_{e,k}$ and depth $D_{e,k}$ at each waypoint, with no simulator object names, object poses, or visibility annotations. 

A segmentation model, SAM2~\citep{sam2_2024}, separates each frame into candidate object instances. We lower the segmenter's confidence threshold to favor over-detection, since a spurious mask can be rejected downstream whereas an object it never proposes cannot be recovered. We filter and group its masks before passing candidates to
the VLM. Each retained mask is then presented to the VLM in three views: a crop of the instance with 15\% padding, the same crop with non-instance pixels masked out, and the full frame with the instance's bounding box drawn. The full frame lets the model use surrounding context to determine what the region contains. The VLM returns a free-text name, such as \emph{vertical storage tank}, and a general category in common words, and a mask is rejected when its category is one of six background terms, such as \emph{ground} or \emph{wall}. Since one kind of object can be named differently across views and episodes, for example \emph{shipping container} and \emph{metal cargo box}, the free-text names on each map are normalized into shared open-vocabulary categories before the critic, the merge, and the questions compare objects by type (Appendix~\ref{app:perception}). We localize each object from its mask and a high depth quantile, then backproject it to world coordinates using the vehicle pose.

Because simulated depth is exact, we add range-dependent depth noise (Appendix~\ref{app:perception}) while leaving pose unperturbed, since pose is the controlled
variable in this study. The resulting object nodes are pooled and clustered
into the initial hierarchical memory $H^{(1)}$, retaining the segmentation, labeling, and localization errors of the perception pipeline for the subsequent method to
identify and repair. Clustering combines spatial proximity with semantic similarity computed using
\texttt{text-embedding-3-small}~\citep{openai2024embeddings} and a
co-visibility constraint requiring at least one shared observing waypoint,
preventing proximity alone from merging objects across occlusions or opposite
approaches.

\subsection{Semantic Critic}

After Episode~1, the robot must diagnose errors from its own memory and observations, $
\mathcal{R}^{(1)} = \mathcal{C}\big(H^{(1)}, Z^{(1)}\big).$ The critic combines two VLM checks, run with \texttt{gpt-5-mini}~\citep{singh2025gpt5} on graph context and the supporting images $Z^{(1)}$, with one statistic computed directly from the graph, and identifies potential errors in $H^{(1)}$ without access to oracle data or external references.

The critic uses three complementary signals: \textbf{plausibility}, which tests whether stored labels are visually and contextually supported; \textbf{density}, which compares visible object estimates with stored objects per waypoint; and \textbf{graph coverage}, a rule on cluster statistics that identifies clusters supported by too few observations or nodes. Together, these target incorrect labels, waypoints with missing objects, and clusters with little supporting evidence. The critic returns plausibility scores and flagged waypoints and clusters, defining the regions $\mathcal{R}^{(1)}$ for Episode~2 to revisit (thresholds in Appendix~\ref{app:metrics}).

\subsection{Corrective Episode and Merge}

The flagged regions from diagnosis must be converted into revisit viewpoints that the robot can physically execute,
$
\mathcal{V}^{(2)} = f_{\mathrm{plan}}(\mathcal{R}^{(1)}, H^{(1)}).
$
\subsubsection{Revisit Planner}
Nearby flagged clusters from the same source waypoint are merged into a single revisit region. For each region, the planner turns in place at the source waypoint, where the mission placed the robot, to face the region centroid. It adds viewpoints $2.5$~m and $4$~m toward the region only when each step still leaves the region's radius plus $2$~m of clearance, and steps $2.5$~m back instead when the region is closer than $6$~m. The distance to the region thus decides which steps a region receives, while the step lengths stay fixed because that distance comes from the observation the critic has just flagged as unreliable. Steps stay on the line to the region, without sideways moves, because that is the only direction whose clearance the planner can check, and the robot carries no obstacle map. Revisits are flown without the Episode~1 degradation, at the capture height and camera pitch of the source waypoint's mission pose, since the degraded pose is the simulated failure and not part of the mission plan. Anchoring revisits to the original viewpoint avoids relying on uncertain region geometry and isolates targeted revisitation from unconstrained viewpoint search. Episode~2 otherwise uses the same motion and recovery procedure as Episode~1 (Appendix~\ref{app:collection-details}), so differences between the episodes come from where the robot looks rather than how it flies.

\subsubsection{Selective Merge}
Episode~2 observations pass through the same perception pipeline and are selectively merged into $H^{(1)}$, thus 
$H^{(2)} = f_{\mathrm{correct}}\big(H^{(1)}, \mathcal{O}^{(2)}\big).$
An incoming detection within $2$~m of a stored object of the same category, seen from a different view, is treated as the same physical object and therefore replaces a stored Episode~1 object or is merged into one already added in Episode~2, while a detection without such a match enters the memory as an object the first pass missed. A stored object is removed only when the critic flagged it as implausible and a revisit had its position in view, unoccluded in depth, without detecting its category there. The $2$~m radius is the one at which the EQA scorer counts detections from different views as one object, no rule depends on how often a type was seen, and objects that no revisit observes stay unchanged.

\section{Evaluation Oracle}
\label{sec:oracle}

Evaluating whether correction improves memory requires a reference for what was actually observable, which is impractical to verify manually at scale. We therefore construct an evaluation oracle from the simulator's scene objects and their ground-truth positions, observed from the same source waypoints at the undegraded camera pose, with the same field of view and range as the perception pipeline, and with occluded objects removed by raycasting. The resulting graph $H^{\mathrm{oracle}}$ is used only for evaluation and is never available to the critic, planner, or selective merge. Simulator asset names often differ from natural object labels, and scene queries also return many non-object actors such as lights, cameras, terrain, and decals. We remove these by name filtering, normalize asset names into plain labels, and pass them through the same open-vocabulary normalization as the memory's names, so that questions and answers refer to one set of categories. $H^{(1)}$ and $H^{\mathrm{oracle}}$ are then compared by region rather than by exact name.

\subsection{Oracle Judge}
The oracle judge provides a common evaluation layer between the perception-built memory and $H^{\mathrm{oracle}}$. Because the two graphs differ in segmentation, localization, and vocabulary, one-to-one object matching would penalize differences that are not memory errors: perception may split one asset into several masks or merge several into one, depth error shifts positions by meters at long range, and a correct label rarely equals an asset name. The judge therefore compares the two graphs over spatial regions rather than individual objects, and treats two labels as matching when their embedding cosine similarity is at least $0.40$ (Appendix~\ref{app:label-agreement}). Within each waypoint's region, it scores memory quality as the mean of three memory-to-oracle ratios: object count, mean nearest-neighbor spacing, and matched object types. It separately evaluates whether $\mathcal{R}^{(1)}$ identifies poorly recovered regions, reporting region-level recall for the critic and for each of its plausibility, density, and coverage signals, with precision and F1 where adequately recovered regions exist. Here a region is a first-level cluster, and it counts as poorly recovered when it holds fewer than half as many objects as the oracle records within $45$~m of it. The same oracle regions provide ground-truth answers for the fixed EQA set.

\begin{figure*}[t]
\centering
\includegraphics[width=\linewidth]{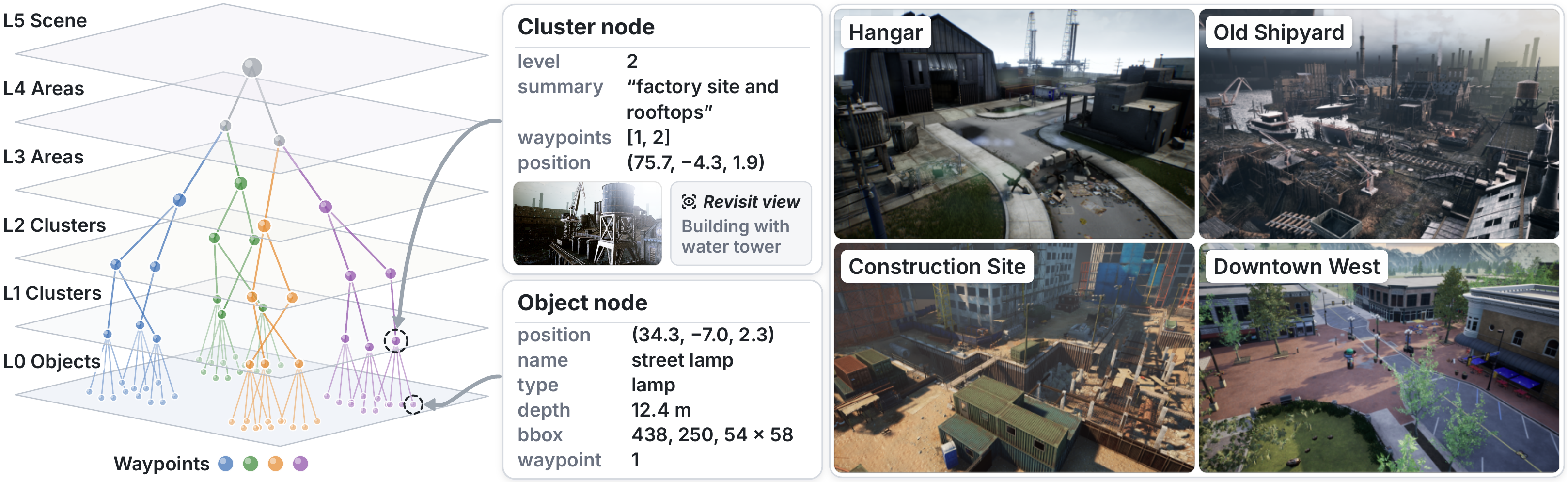}
\caption{\textbf{Hierarchical memory and evaluation environments.} Left:
objects are grouped into clusters and clusters into areas, as in
Embodied-RAG, and the first cluster level is the unit the critic flags. Middle: example cluster and object records. A corrective
episode acts at the object tier, removing flagged entries that a revisit
looked at and did not find, replacing entries it observes again, and adding
ones it newly recovers, after which the affected clusters are rebuilt. Right: the four outdoor maps spanning an enclosed courtyard, an open industrial yard, a construction area, and a downtown plaza.}
\label{fig:memory}
\end{figure*}

\section{Experiments and Evaluation}
\label{sec:experiments}

\subsection{Simulation Platform and Scope}

All experiments run in AirSim \citep{airsim2017} on adapted Unreal Engine~4
environments using the Docker setup released with Embodied-RAG \citep{embodiedrag2024}. Unlike Embodied-RAG's released AirSim pipeline, which builds memory from simulator ground-truth object names and poses,
our perception pipeline builds memory from RGB-D observations while a separate oracle pipeline reads simulator state only for evaluation, both at $1280\times720$ resolution and $90^\circ$ horizontal field of view. The study isolates perception and memory correction by treating \emph{(i)} pose
as exact, excluding localization error; \emph{(ii)} simulator depth as exact
before the sensor model is applied, making that model the controlled source of
range error; and \emph{(iii)} the environment as static across episodes. Evaluation is restricted to outdoor environments because revisit poses must be reachable without an occupancy map or obstacle-avoidance stack, which are outside the present scope; open outdoor scenes leave room to move around each source waypoint without collision planning. 

\subsection{Maps, Viewpoints, and Viewpoint Conditions}
\label{sec:maps-conditions}

The evaluation uses the four outdoor maps shown in Figure~\ref{fig:memory}:
Hangar, Old~Shipyard, Construction~Site, and Downtown~West. Each map has four
hand-selected source waypoints chosen for object diversity, density, and
regional isolation, and Episode~1 follows the same fixed source locations in
every run. One observation is recorded at each waypoint and none in transit; later episodes visit only the revisit waypoints generated by the planner.

Each map is evaluated under the undegraded condition and two degraded viewpoint conditions. \emph{Undegraded} uses the source poses unchanged; \emph{pitch-up} tilts the
camera upward, removing object-bearing near-field content; and
\emph{low-lateral} lowers the camera close to the ground and turns it $45^\circ$ off its intended heading (exact values in Appendix~\ref{app:collection-details}). The degradations make the source of error controlled, since without them Episode~1 errors would be whatever perception happened to produce and a gain could not be traced to recovering anything. The oracle always observes the undegraded poses, so it records the content a degraded run loses and thereby defines what correction should recover, while the critic and planner never see the degradation.

\subsection{Metrics}

Critic and memory quality are scored by the oracle judge described in Sec.~\ref{sec:oracle}, with metric definitions and critic thresholds given in Appendix~\ref{app:metrics}. Our primary metric is fixed-set EQA, which evaluates $H^{(1)}$ and $H^{(2)}$ on identical questions to measure how correction changes the KB. Each question asks about the region seen from one source waypoint, so a loss at that waypoint shows up in its own questions instead of being averaged over the map. The question set covers four families (inventory, dominant type, type richness, and rare presence) to test whether correction recovers different forms of scene information.

A question's region is a circle centered on the mean position of the oracle objects visible from its waypoint, with a radius equal to their largest offset from that center along either ground axis and at least $15$~m. The ground truth counts the oracle objects inside it, and the answers from $H^{(1)}$ and $H^{(2)}$ apply one fixed counting rule to the stored objects inside it, whichever viewpoint observed them, counting detections of one category within $2$~m from different views once, so the LLM judge only compares answers. Questions use only categories present in both the oracle and the map's undegraded reference memory, so every question is answerable by a perception-built memory. An LLM judge (\texttt{gpt-5-mini}) with a fixed rubric scores answers as correct ($1.0$), partial ($0.5$), or wrong ($0.0$), and count questions are answered in ranges (none, one, 2--3, 4--6, 7--10, 11--20, more than 20), with an answer one range from the reference receiving partial credit. Questions that fail to score because of a judge or retrieval error are excluded from the mean and reported separately. Question-family definitions, count ranges, and the full judging rubric are given in Appendix~\ref{app:metrics}.

\subsection{Controls and Ablations}

The main Episode~1-to-Episode~2 evaluation includes three ablations that isolate each component of SPARK: \emph{(i)} the merge ablation compares unchanged $H^{(1)}$, append-only updating, cluster replacement, and the selective merge; \emph{(ii)} the critic ablation evaluates plausibility, density, and coverage individually and together; and \emph{(iii)} the revisit ablation compares critic-guided and untargeted placement under the same Episode~1 memory, observation budget, and merge. All four maps are evaluated and averaged, because a single map holds only four source waypoints and its result can move with a few questions. Unreachable revisit candidates are skipped and reported so failed proposals remain part of the planner evaluation.

We do not adapt prior systems as baselines because each either does not apply to embodied memory or reduces to one of our controls. CRAG corrects retrieval with external text sources such as web search, and no such source exists for what a robot observed in its own environment, where re-observation is the only corrective evidence. Mind Palace, without its question-conditioned planner and temporal objective, reduces to re-observing the scene and appending each observation to memory, which is the append-only condition of the merge ablation.

\section{Results}
\label{sec:results}

We first report the Episode~1 to Episode~2 transition across four maps and two degraded viewpoint conditions, then use the four-map ablations to isolate the merge policy, the critic signals, and the revisit policy, and lastly follow correction through Episodes~3 and~4. Each EQA comparison uses the same fixed question set before and after the relevant intervention.

\subsection{Main Results: Episode~1 to Episode~2}

\begin{table}[htbp]
\centering
\setlength{\abovecaptionskip}{4pt}
\setlength{\belowcaptionskip}{4pt}
\begin{minipage}[t]{0.52\linewidth}
\centering
\caption{Mean EQA over the four maps before and after one corrective episode, with the undegraded reference from the same pipeline.}
\label{tab:main}
\small
\begin{tabular}{lccc}
\toprule
Viewpoint condition & $H^{(1)}$ & $H^{(2)}$ & $\Delta$ \\
\midrule
Pitch-up     & 0.231 & 0.526 & $+0.295$ \\
Low-lateral  & 0.381 & 0.535 & $+0.154$ \\
\midrule
Overall      & 0.306 & 0.531 & $+0.225$ \\
\midrule
\textit{Undegraded ref.} & \multicolumn{2}{c}{\textit{0.488}} & --- \\
\bottomrule
\end{tabular}
\end{minipage}\hfill
\begin{minipage}[t]{0.44\linewidth}
\centering
\caption{Merge ablation on the same pitch-up Episode~2 observations, with mean objects per map.}
\label{tab:merge}
\small
\begin{tabular}{lcc}
\toprule
Merge policy & $\Delta$ EQA & Objects \\
\midrule
Unchanged $H^{(1)}$  & $+0.000$ & --- \\
Append only          & $+0.296$ & 391 \\
Replace clusters     & $+0.294$ & 370 \\
Selective merge      & $+0.295$ & 283 \\
\bottomrule
\end{tabular}
\end{minipage}
\end{table}

Correction improves EQA across all four maps and both viewpoint conditions. Some recovery is expected because Episode~2 views the flagged regions without the degradation, but it still requires the critic to locate the losses from the memory alone, without seeing the degradation or the oracle, and the revisits and merge to turn new views into correct answers without losing existing ones. Pitch-up loses whole categories, leaving Episode~1 memories of $10$ to $76$ objects per map against $85$ to $103$ for low-lateral, and therefore has more to recover ($0.231$ to $0.526$, against $0.381$ to $0.535$). Overall, one corrective episode of $6$ to $23$ captured views per map raises EQA from $0.306$ to $0.531$, above the $0.488$ undegraded reference, with 13--41\% fewer stored objects than append-only updating (Sec.~\ref{sec:merge-abl}). The reference is a single pass with one pose per waypoint, whereas the corrected memory adds revisit views from further poses, and three of the four maps exceed their reference after one correction, with Construction~Site below it in both conditions. On Hangar, whose source poses face shadowed areas, even the low-lateral Episode~1 memory scores above the reference ($0.470$ against $0.342$). The largest gain is on Old~Shipyard under pitch-up, where EQA rises from $0.233$ to $0.629$, and the smallest on Hangar under low-lateral ($+0.017$), where revisits fix answers at one waypoint but lose counts at others.

How much correction helps depends on what the first pass missed. If a degraded view misses the containers that dominate a waypoint, one revisit that observes them can correct both its dominant-type and type-richness answers, whereas ten more instances of a type already stored change an inventory answer only if the count crosses into the next range. Memory size therefore grows far more than EQA, as when the Old~Shipyard pitch-up memory goes from $20$ to $290$ objects. Our maps repeat similar categories across waypoints, so many added objects leave the answers unchanged, and the gains reflect these maps, degradations, and questions, not an upper bound on correction.

\subsection{Merge Ablation}
\label{sec:merge-abl}

All three active rules raise EQA by about the same amount, from $+0.294$ to $+0.296$ (Table~\ref{tab:merge}), so the gain comes from adding the Episode~2 evidence and not from the choice of rule. The selective merge reaches this gain while storing 13--41\% fewer objects per map than append (283 against 391 on average, and 370 for replacement), because it keeps repeated detections of one object only once, a saving that matters as memory and query cost grow with every episode (Sec.~\ref{sec:longitudinal}). The three rules score alike because the pitch-up memories hold only $10$ to $76$ objects before correction, which leaves little Episode~1 content for append to keep in error or for replacement to discard. In a larger memory, or over more episodes, append would keep every earlier error and replacement would drop valid objects in each revisited cluster, which one pitch-up transition does not test.

\subsection{Critic Ablation}
\label{sec:controls}

Revisits target only the regions the critic flags, so we test whether the three signals flag different regions by running plausibility, density, and graph coverage individually and together, scoring each flagged set against the regions the oracle shows to be poorly recovered.

\begin{table}[htbp]
\centering
\setlength{\abovecaptionskip}{4pt}
\setlength{\belowcaptionskip}{4pt}
\begin{minipage}[t]{0.48\linewidth}
\centering
\caption{Critic ablation over all four maps, region-level recall (precision undefined, Appendix~\ref{app:metrics}).}
\label{tab:critic}
\small
\begin{tabular}{lccc}
\toprule
Diagnosis & Recall & TP & FN \\
\midrule
Plausibility only & $0.267$ & 16 & 44 \\
Density only      & $0.117$ & 7 & 53 \\
Coverage only     & $0.667$ & 40 & 20 \\
\textbf{All three} & $\mathbf{0.800}$ & 48 & 12 \\
\bottomrule
\end{tabular}
\end{minipage}\hfill
\begin{minipage}[t]{0.48\linewidth}
\centering
\caption{EQA gain of critic-guided and untargeted revisits on the low-lateral memories, same budget.}
\label{tab:revisit}
\small
\setlength{\tabcolsep}{3.5pt}
\begin{tabular}{lcc}
\toprule
Map & Critic-guided & Untargeted \\
\midrule
Old~Shipyard      & $+0.300$ & $+0.400$ \\
Hangar            & $+0.017$ & $-0.022$ \\
Construction~Site & $+0.167$ & $+0.138$ \\
Downtown~West     & $+0.134$ & $+0.071$ \\
\midrule
Mean              & $\mathbf{+0.154}$ & $+0.147$ \\
\bottomrule
\end{tabular}
\end{minipage}
\end{table}

Table~\ref{tab:critic} shows that coverage alone finds the most poorly recovered regions, since it identifies regions with little supporting evidence, while plausibility finds about a quarter and density, which flags whole waypoints, about a tenth. Combining all three raises recall from $0.667$ to $0.800$, so the signals capture different failures and further independent checks could raise it again, while the $12$ regions the combined critic misses bound what one round of revisits can repair.

\subsection{Revisit Ablation}
\label{sec:revisit-abl}

After diagnosis, the system must decide where to spend its revisit budget. We compare the critic-guided planner with an untargeted baseline that uses the same Episode~1 memory, planner, capture heights and camera pitch, number of viewpoints, and selective merge, and differs only in which first-level clusters it revisits. It takes the clusters the critic did not flag first, then flagged ones, each in an order fixed by one random seed, until the planner has produced as many viewpoints as the guided plan, because a uniform draw would mostly repeat the guided choice when the critic flags $62$ to $81\%$ of the clusters of a degraded low-lateral memory.

Table~\ref{tab:revisit} shows that critic-guided revisitation yields a larger mean gain than untargeted revisitation, $+0.154$ against $+0.147$, with about as many views flown ($58$ of $77$ candidates, against $56$ of $76$). The margin is small because the untargeted baseline is a strong control by construction: it flies the same planner and budget around the same source waypoints, and once its $6$ to $15$ unflagged clusters per map run out, the rest of its budget goes to flagged ones. The guided arm leads on three of the four maps. On Old~Shipyard, most of the difference comes from one waypoint, where the guided view, aimed at the centroid of five merged flagged clusters, left out of frame a stack of barrels that an untargeted view aimed at one of those clusters captured. Its advantage should grow in later episodes, when fewer regions remain flagged and the choice of where to look matters more.

\subsection{Longitudinal Study: Episode~1 to Episode~4}
\label{sec:longitudinal}

We extend the correction loop to Episode~4 on all four maps, with each episode diagnosing and revisiting the memory produced by the previous one. Episode~2 is the low-lateral main result, also the critic-guided arm of the revisit ablation. The study uses the low-lateral condition on every map because its Episode~1 performance is closer to the undegraded reference than pitch-up.

\begin{figure}[htbp]
\centering
\setlength{\abovecaptionskip}{4pt}
\setlength{\belowcaptionskip}{4pt}
\includegraphics[width=0.65\linewidth]{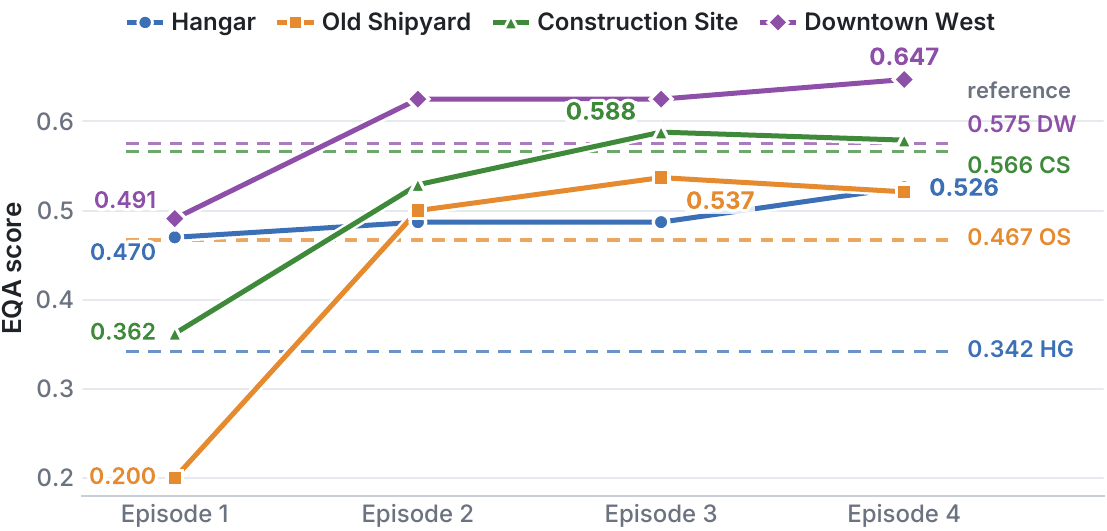}
\caption{Fixed-set EQA across four episodes, with each map's undegraded reference as a dashed line.}
\label{fig:longitudinal}
\end{figure}

Figure~\ref{fig:longitudinal} shows that all four maps exceed their undegraded reference by Episode~3, because that reference is a single episode with one pose per waypoint, and additional views recover objects that were occluded, merged, or poorly observed even from the original viewpoints. Old~Shipyard gains most in the first correction ($0.200$ to $0.500$); Construction~Site crosses its reference at Episode~3 ($0.588$ against $0.566$); Hangar starts above its reference and gains mainly at Episode~4; and Downtown~West exceeds its reference after one correction and rises slightly to $0.647$ as its memory grows from $467$ to $1{,}142$ objects. Gains flatten after Episode~3, so the useful number of revisits depends on how much recoverable information remains, and a loop could stop once a further episode no longer changes the answers.

Repeated correction also exposes a scaling cost that single-episode evaluations do not show: each episode adds observations, so the KB grows in size and query cost even though the selective merge stores fewer objects than append. By Episode~4, Old~Shipyard stores $958$ objects for a scene with $152$ oracle objects, and Downtown~West $1{,}142$ for $493$, even though each corrective episode is capped at thirty viewpoints. Long-range depth error places repeated observations of one object farther apart than the $2$~m the selective merge treats as one object, so sustaining the loop requires cross-episode association that tolerates such error.

\section{Conclusion}

We presented SPARK, a corrective memory loop in which a robot diagnoses its KB without ground truth, revisits flagged regions, and selectively adds, removes, or updates what it stores. Across four outdoor scenes and two degraded viewpoint conditions, a corrective episode improves downstream EQA on every map, exceeding an undegraded reference on average, and critic-guided revisitation yields a larger mean gain than an untargeted policy under the same budget.

Our experiments isolate memory correction in a static world with known pose and a fixed motion routine. One view per flagged cluster, or sideways moves once an obstacle map makes them safe, would keep every cluster of a wide region in frame. Longer deployments will require persistent object association to merge repeated observations and control KB growth. The same capability could help distinguish scene changes from perception errors and account for pose drift under real SLAM. Extending correction online during a mission and testing across indoor environments and other platforms would further expose different scene geometry and sensing constraints.

\subsection*{AI use statement}

We used generative AI tools to assist with language editing, including polishing, shortening, and clarifying prose and LaTeX formatting. We did not use generative AI for research ideation, experimental design, or interpretation of results. All technical content, claims, and conclusions were produced by the authors. We take responsibility for the final content of this work.

\bibliography{iclr2027_conference}
\bibliographystyle{iclr2027_conference}

\clearpage
\appendix

\section{Correction Loop}
\label{app:algorithm}

This appendix presents the full loop and the thresholds used at each stage, so the path from a captured frame to a corrected memory can be followed end to end. Algorithms~\ref{alg:spark}--\ref{alg:spark-correct} split the procedure into three parts: Episode~1 collection and memory construction, diagnosis and revisit planning, and the corrective episode with its evaluation. The procedure is the same across viewpoint conditions, with the condition affecting only the poses produced by \textsc{ApplyCondition} in Stage~1 and remaining hidden from all decision-making components.

\begin{algorithm}[t]
\caption{SPARK correction loop, Part~1: Episode~1 collection and memory
construction.}
\label{alg:spark}
\begin{algorithmic}[1]
\Require waypoints $W = \{w_1,\dots,w_4\}$, viewpoint condition $c$,
depth limit $R = 40$~m
\Ensure corrected memory $H^{(2)}$, EQA scores on $H^{(1)}$ and $H^{(2)}$

\Statex \textbf{\textsc{Stage 1: Episode 1 collection}}
\State $P \gets \textsc{ApplyCondition}(W, c)$
\Comment{undegraded, pitch-up ($+40.7^\circ$), or low-lateral}
\For{each pose $p \in P$}
  \State climb to $20$~m, translate, descend, set yaw on airframe and
         pitch on gimbal, stabilize
  \State record $(I_t, D_t, \mathbf{p}_t)$
  \Comment{one frame per waypoint; nothing recorded in transit}
\EndFor

\Statex \textbf{\textsc{Stage 2: Build $H^{(1)}$ from RGB-D only}}
\For{each observation $t$}
  \State $M \gets \textsc{Segment}(I_t)$
         \Comment{IoU $0.50$, stability $0.80$, $32\times32$ grid}
  \State discard masks with area $<300$~px$^2$ or dimension $<20$~px
  \State $D_t \gets \textsc{AddSensorNoise}(D_t)$
         \Comment{$\sigma_z = 0.002z^2$, $1.2$~px jitter, $2\%$ dropout
                  on grazing surfaces}
  \State $z_m \gets Q_{75}$ of $D_t$ over each mask $m$;
         discard if $z_m > R$
  \State $C \gets$ 15 largest masks $\cup$ up to 5 groups of small
         masks clustered at $6\%$ of the image diagonal
  \For{each candidate $m \in C$}
    \State $(n_m, c_m) \gets \textsc{Caption}(\text{crop},
           \text{masked crop}, \text{boxed frame})$
           \Comment{free-text name, general category}
    \State \textbf{reject} if $c_m$ is a background term or its head
           noun is a background word; caption again without \emph{person}
           if a person fails verification
    \State backproject $z_m$ through the mask centroid to world
           coordinates
  \EndFor
\EndFor
\State pool all objects; cluster by complete linkage at level-$\ell$
       threshold $5.0(1+\ell)$~m, blocking merges that violate the
       semantic or co-visibility constraint
\State $H^{(1)} \gets$ resulting hierarchy;
       $Z^{(1)} \gets$ retained images and regional summaries
\State type every object by the open-vocabulary category of $n_m$
       \Comment{one mapping per map, Appendix~\ref{app:perception}}

\end{algorithmic}
\end{algorithm}

\begin{algorithm}[t]
\caption{SPARK correction loop, Part~2: diagnosis and revisit planning. Neither
stage reads the oracle.}
\label{alg:spark-diagnosis}
\begin{algorithmic}[1]
\Statex \textbf{\textsc{Stage 3: Critic}}
\State $S \gets \textsc{SceneSummary}(Z^{(1)})$
\For{each stored object $o$}
  \State $\pi_o \gets \textsc{Plausibility}(o, Z^{(1)}, S) \in [0,1]$;
         \textbf{flag} if $\pi_o < 0.6$ or unscored
\EndFor
\For{each waypoint $w$}
  \State $\rho_w \gets |H^{(1)}_w| \,/\, \textsc{EstimateVisible}(I_w)$;
         \textbf{flag} if $\rho_w < 0.4$
  \Statex \hskip\algorithmicindent
         \Comment{zero estimate: unassessable in the main results,}
  \Statex \hskip\algorithmicindent
         \Comment{flagged if $|H^{(1)}_w| > 0$ in the critic ablation}
\EndFor
\For{each level-1 cluster $k$}
  \State \textbf{flag} if $\mathrm{support}(k) < 2$
         \textbf{and} $\big(|k| \le 2$ \textbf{or}
         $\bar{d}(k) \ge 40\text{~m}\big)$
\EndFor
\State $\mathcal{R}^{(1)} \gets$ union of the three flagged sets

\Statex \textbf{\textsc{Stage 4: Revisit planning}}
\State merge flagged clusters within $25$~m that share a source waypoint
\For{each region $r \in \mathcal{R}^{(1)}$}
  \State $g \gets \|\mathbf{x}_{\text{drone}} - \mathbf{x}_r\|_{xy}$;
         $c \gets \max(\text{radius}(r) + 2, 4)$~m
  \State emit \textsc{Rotate}: stay at the source waypoint
  \If{$g > c + 2.5$~m} emit \textsc{Forward} $2.5$~m, and $4$~m if
       $g > c + 4$~m
  \ElsIf{$g < 6$~m} emit \textsc{Back} $2.5$~m
  \EndIf
  \Statex \hskip\algorithmicindent
         \Comment{steps run along the bearing to $r$; no sideways moves}
  \State set altitude and camera pitch to those of the source waypoint
  \Statex \hskip\algorithmicindent
         \Comment{mission pose, without the Episode 1 degradation}
\EndFor
\State $\mathcal{V}^{(2)} \gets$ emitted candidates, each yawed to face
       the region centroid

\end{algorithmic}
\end{algorithm}

\begin{algorithm}[t]
\caption{SPARK correction loop, Part~3: corrective episode, merge, and evaluation.}
\label{alg:spark-correct}
\begin{algorithmic}[1]
\Statex \textbf{\textsc{Stage 5: Episode 2 and selective merge}}
\State fly $\mathcal{V}^{(2)}$ with the Stage-1 motion routine; skip
       unreachable candidates and record them as skipped
\State build incoming objects with the Stage-2 pipeline
\For{each incoming object $o$}
  \State $s \gets$ nearest object of the same category within $2$~m,
         seen from a different view
  \If{$s$ is stored in $H^{(1)}$} \textbf{replace} $s$ by $o$
  \ElsIf{$s$ was added in Episode 2} \textbf{merge} $o$ into $s$
  \Else\ \textbf{add} $o$
  \EndIf
\EndFor
\For{each stored object flagged by plausibility}
  \State \textbf{remove} if a revisit frame had its position in view,
         unoccluded in depth, and no incoming object of its category
         lies within $2$~m
\EndFor
\State $H^{(2)} \gets$ hierarchy rebuilt from the merged objects with the
       Stage-2 clustering

\Statex \textbf{\textsc{Stage 6: Evaluation (oracle, never used by Stages 1--5)}}
\State collect the oracle at \emph{undegraded} poses under the same
       FOV and depth limit, with ray-cast occlusion
\State score critic quality as region-level recall (precision where defined)
\State generate the fixed question set from oracle-supported regions
\State score the same questions on $H^{(1)}$ and $H^{(2)}$;
       report the paired difference
\end{algorithmic}
\end{algorithm}

The critic only decides which regions need a revisit and leaves the viewpoints to the planner, which places them from the region's geometry and the source waypoint; hence a revisit still helps when the critic finds an unreliable region but misjudges what occurred there. Evaluation is not part of the deployment loop, as the critic reads only $H^{(1)}$ and $Z^{(1)}$, and the oracle is collected at Stage~6, after $H^{(2)}$ is fixed, so no oracle information can shape the corrected memory.

Each merge operation is triggered by a specific detection, or by its absence from a frame that had the object in view, and never by how often a type was observed. Two detections of one category within $2$~m from different views are treated as one object because that is the radius at which the EQA scorer counts them once, so the merge stores what the evaluation would count once and cannot gain by storing duplicates. A re-observed stored object is replaced rather than kept beside the new detection, because the revisit views it from the mission pose without the degradation. Removal needs two pieces of evidence, a plausibility flag and a revisit frame that had the object's position in view without detecting its category there, because a flag alone can reflect an uncertain caption and an absent detection means little unless the camera saw that spot. A frequency threshold is not used, because objects seen once are often the rare objects a degraded view loses first, and discarding them would remove correct content along with noise. The in-view test projects the stored position into each revisit frame with the same camera model and pose convention used for backprojection (Appendix~\ref{app:perception}), so a stored object maps back to the pixel and depth it came from.

\section{Perception Pipeline}
\label{app:perception}

\subsection{Segmentation and candidate selection}

An object the segmenter does not propose cannot be utilized later, although surplus masks can still be rejected at captioning, hence segmentation is tuned for recall. Thus, we run class-agnostic SAM2~\citep{sam2_2024}, $\mathcal{M}_{e,k} = \mathrm{SAM}(I_{e,k})$, which prompts the model with a $32\times32$ grid of points instead of clicks or category names, and lower its predicted-IoU and stability thresholds to $0.50$ and $0.80$, below their defaults. Candidates are then filtered in the order of Algorithm~\ref{alg:spark}: masks smaller than $300$~px$^2$ or narrower than $20$~px are removed as likely texture fragments, and masks beyond the $40$~m depth limit are discarded once their depth is known. Since every remaining candidate costs one VLM call, the rest are ranked for a captioning budget that keeps the 15 largest masks per waypoint and up to five groups of smaller masks, formed by clustering their positions at $6\%$ of the image diagonal so that small objects lying close together are still represented.

\subsection{Captioning and object rejection}

\begin{figure}[t!]
\centering
\includegraphics[width=\linewidth]{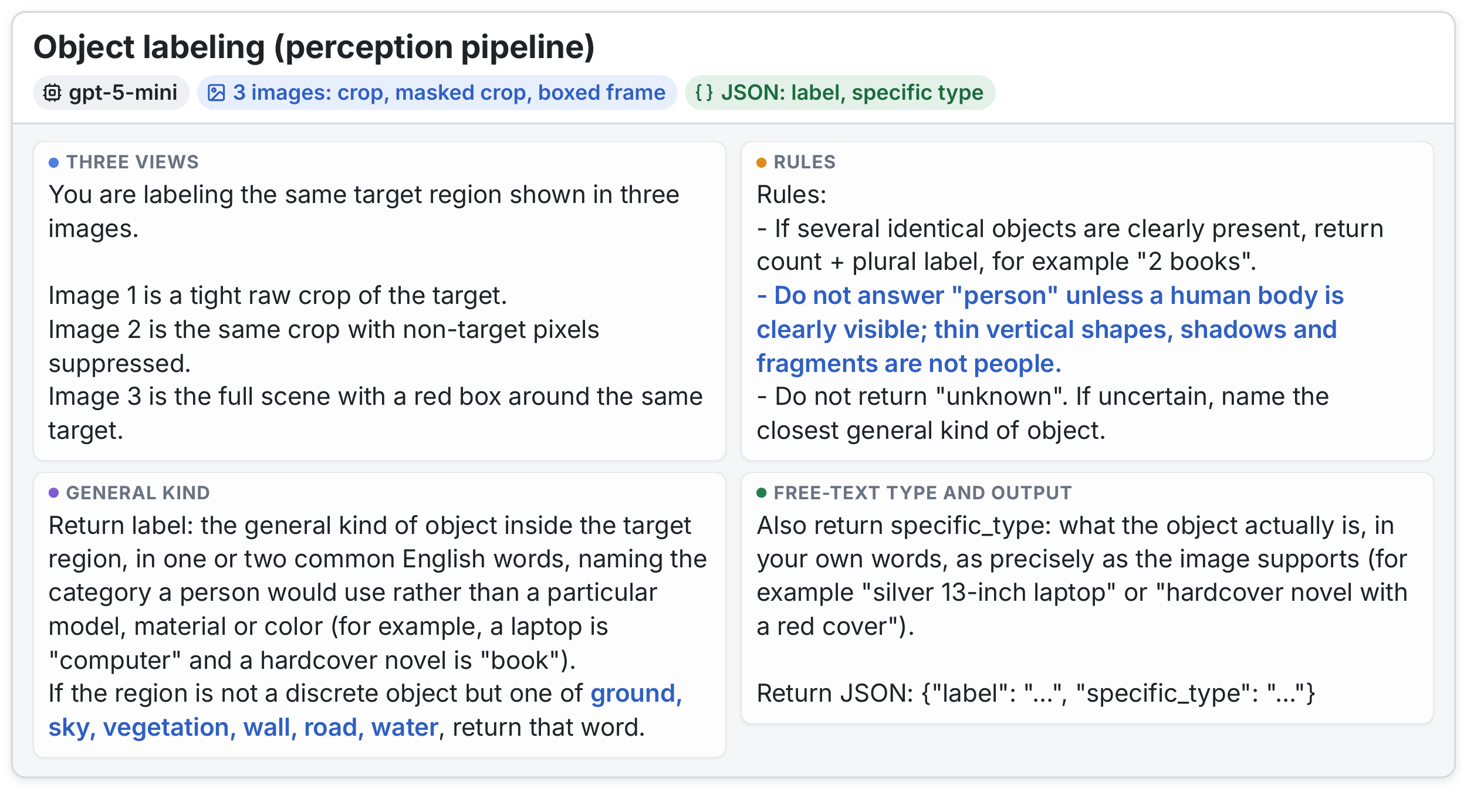}
\caption{Object-labeling prompt. Each retained mask is sent in one call with three images: the padded crop, the masked crop, and the full frame with the bounding box drawn. A label equal to one of the six background terms (\emph{ground} to \emph{water}) rejects the mask, and the free-text \emph{specific type} is the name $n$ that the memory stores. The person rule shown is the one in the first call, and a label that fails person verification is requested again with \emph{person} excluded. Prompt text in this and the following prompt figures is copied from the code that produced the reported results: blue placeholders are filled at run time, \textnormal{[\ldots]} marks text omitted for length, and the small headings are added for readability.}
\label{fig:prompt-perception}
\end{figure}

A single tight crop loses the context of surrounding objects, and the full image, even with the region highlighted, is often not enough, which is why each captured mask $m$ is sent to \texttt{gpt-5-mini} in a single call with three views, a crop with 15\% padding, the same crop with non-target pixels masked out, and the full frame with the candidate's bounding box drawn (Figure~\ref{fig:prompt-perception}),
\begin{equation}
(n,c)=\mathrm{VLM}(m_{\mathrm{crop}},
                   m_{\mathrm{isolated}},
                   m_{\mathrm{context}}),
\end{equation}
where $n$ is a free-text name for the object, such as \emph{vertical storage tank}, and $c$ is its general kind in common words, or one of six background terms when the region is not a discrete object. The prompt gives no list of object types, and its two examples of a general kind (a laptop is a \emph{computer}, a hardcover novel a \emph{book}) come from outside the evaluated scenes. The memory stores $n$, and $c$ serves only to reject masks that are not objects. Because the permissive segmenter also proposes surface fragments such as pavement and walls, a proposal is rejected when $c$ is a background term or its head noun is a background word, and a predicted \emph{person} that fails a verification call is captioned again with \emph{person} excluded. Without this step, such fragments would enter memory as generic objects and inflate later counts. Background matching uses whole words; consequently, compounds such as \emph{roadblock} and \emph{sandbag} are kept.

\subsection{Open-vocabulary categories}

The memory, the critic, the merge, and the EQA questions all count and compare objects by type, for example how many containers were visible from a waypoint or which type was most common there. Free-text names vary with viewpoint and wording, so the same object can be stored as \emph{shipping container} in one view and as \emph{metal cargo box} in another, and counts, merges, and answers would then depend on wording rather than on perception. We therefore normalize names into categories once per map, using two \texttt{gpt-5-mini} calls and no predefined list. The first reduces every distinct name to the head noun for its kind of object, one or two common words without material, color, size, count, state, or position, so \emph{stacked wooden slatted storage crates} becomes \emph{crate} and the asset name \emph{sm tank vertical} becomes \emph{tank}, and it marks names that describe no discrete object, such as ground or a wall surface. The second groups head nouns that name the same kind of object and gives each group its most common everyday word; it runs three times, and each noun keeps the category it receives most often. Across all episodes reported on each map, this reduces $2{,}055$ to $4{,}084$ distinct names to $144$ to $220$ categories.

The same mapping assigns categories in $H^{(1)}$, $H^{(2)}$, the undegraded reference, and the oracle of a map, so normalization cannot by itself create a gain from correction. A name first seen in a later episode is reduced to its head noun in the same way and joins the category that already holds that noun, or starts a new one, leaving every earlier assignment unchanged. The grouping call sees the oracle's asset nouns alongside the memory's so that questions and answers share categories, yet this does not shape how memory objects are grouped: regrouping the memory nouns alone agrees with the reported grouping on $97.8$ to $99.8\%$ of object pairs across the four maps, against $97.9$ to $99.4\%$ for a plain rerun that includes the oracle nouns. Normalization does not remove perception errors, because masks are still split or merged, the model can still misname an object, and positions still carry depth error.

\subsection{Depth, backprojection, and sensor noise}

To place each object in 3D, we need a single depth for its surface rather than for the background behind it or an occluder at its edge. We take the $75$th percentile of the depth values inside the mask,
\begin{equation}
d = Q_{75}\big(\{\, D_{e,k}(p) : p \in m \,\}\big),
\end{equation}
instead of sampling the bounding-box center, which can fall on background for thin or irregular objects. Using the whole mask limits background contamination, and a percentile above the median favors the object's surface over foreground pixels near the mask boundary.
Backprojection converts each object to world coordinates with the vehicle pose and AirSim's NED convention \citep{airsimdocs}, treating the depth value as depth along the optical axis, although the depth image stores AirSim's perspective depth, the range along each pixel's viewing ray. An object at angle $\alpha$ from the optical axis is therefore placed farther along its viewing ray by a factor of $1/\cos\alpha$, about $6\%$ at $20^\circ$ and up to $41\%$ at the left and right image edges. The same backprojection is used in every episode and viewpoint condition; accordingly, this error is present in both $H^{(1)}$ and $H^{(2)}$ and adds to the localization error that the critic and merge must tolerate.

A memory built from exact simulator depth would contain none of the range errors of a real sensor, so we add noise whose standard deviation grows with the square of range, the form reported for structured-light depth sensors \citep{khoshelham2012}, $\sigma_z = kz^2$. We set $k = 0.002~\mathrm{m}^{-1}$, which gives an error of about $2\%$ of range at $10$~m and $8\%$ at $40$~m. This applies the quadratic form well beyond the few meters in which it was measured, so the noise should be read as a deliberately harsh setting rather than a calibrated sensor model. \citet{nguyen2012} find depth noise nearly constant until the angle between the surface normal and the viewing direction reaches about $60$--$70^\circ$, after which it rises sharply along with the fraction of invalid pixels, and we follow that trend by increasing noise and pixel dropout on surfaces seen at grazing angles, with a lateral jitter of $1.2$~px. Pose drift is not added, which keeps depth error separate from pose, the controlled variable in this study.

\subsection{Models}

Segmentation runs SAM2~\citep{sam2_2024} on a hosted endpoint, and every language and vision call uses \texttt{gpt-5-mini} for captioning and validity judgement, the critic, cluster naming and summaries, and the EQA judge. We chose this small multimodal model because its low cost and latency allowed the many repeated collections and scoring passes this study required, while remaining capable enough for labeling and judging. We refer to calls with image inputs as VLM calls and to text-only calls as LLM calls.

\subsection{Clustering}

Clustering decides which detections the memory treats as one region, and the first level of the resulting hierarchy is the unit that the critic flags and the evaluation scores. Detections are pooled across waypoints and grouped by complete-linkage agglomeration with the level-dependent threshold in Algorithm~\ref{alg:spark}. Embodied-RAG describes clustering by a weighted combination of spatial and semantic similarity, but its released implementation passes only positions to the clustering algorithm and adds semantics afterward through cluster summaries, so we restore semantic similarity as a constraint on merging and add a co-visibility constraint.

To keep different object types in separate clusters, the semantic constraint compares labels by cosine similarity of \texttt{text-embedding-3-small} embeddings at a threshold of $0.40$, because exact string matching would treat descriptions such as \emph{corrugated metal fence panel} and \emph{metal panel perimeter fence}, which name the same type of object, as different. Embeddings for all graph labels are computed in one call, and a pair without a similarity is allowed to merge; therefore, a missing embedding cannot silently split a cluster.

The co-visibility constraint addresses a failure of purely spatial clustering, which can join nearby objects that no single waypoint saw together, such as objects on opposite sides of a wall or approached from different directions. Without the constraint, objects from a flagged waypoint can also be absorbed into clusters anchored elsewhere, which leaves that waypoint without a revisit region. A merge is therefore blocked when no waypoint observed both objects. The rule is symmetric, so the result does not depend on processing order, and it applies only when both nodes carry waypoint provenance, leaving objects without recorded provenance to be clustered by distance alone. Clusters above the first level do not keep provenance, so the constraint acts at the first level and in the final pass that reattaches remaining unclustered nodes to nearby clusters by distance.

We do not present co-visibility as a separate contribution, because measuring its effect on EQA would require dedicated experiments that cannot reuse the infrastructure of this study. We added it because purely spatial-semantic clustering can cause the failures described above, and it has a measurable effect outside EQA: on Hangar, co-visibility reduced cross-waypoint clusters from two to zero without changing the total number of clusters.

\subsection{Resulting error modes}

The resulting memory keeps the errors a real perception pipeline makes, such as adjacent instances merged into one, single objects split across masks, vague or wrong names for small or partially occluded objects, and localization errors of several meters from single-frame backprojection.

\section{Metric Definitions}
\label{app:metrics}

\subsection{Critic thresholds}

\begin{figure}[t!]
\centering
\includegraphics[width=\linewidth]{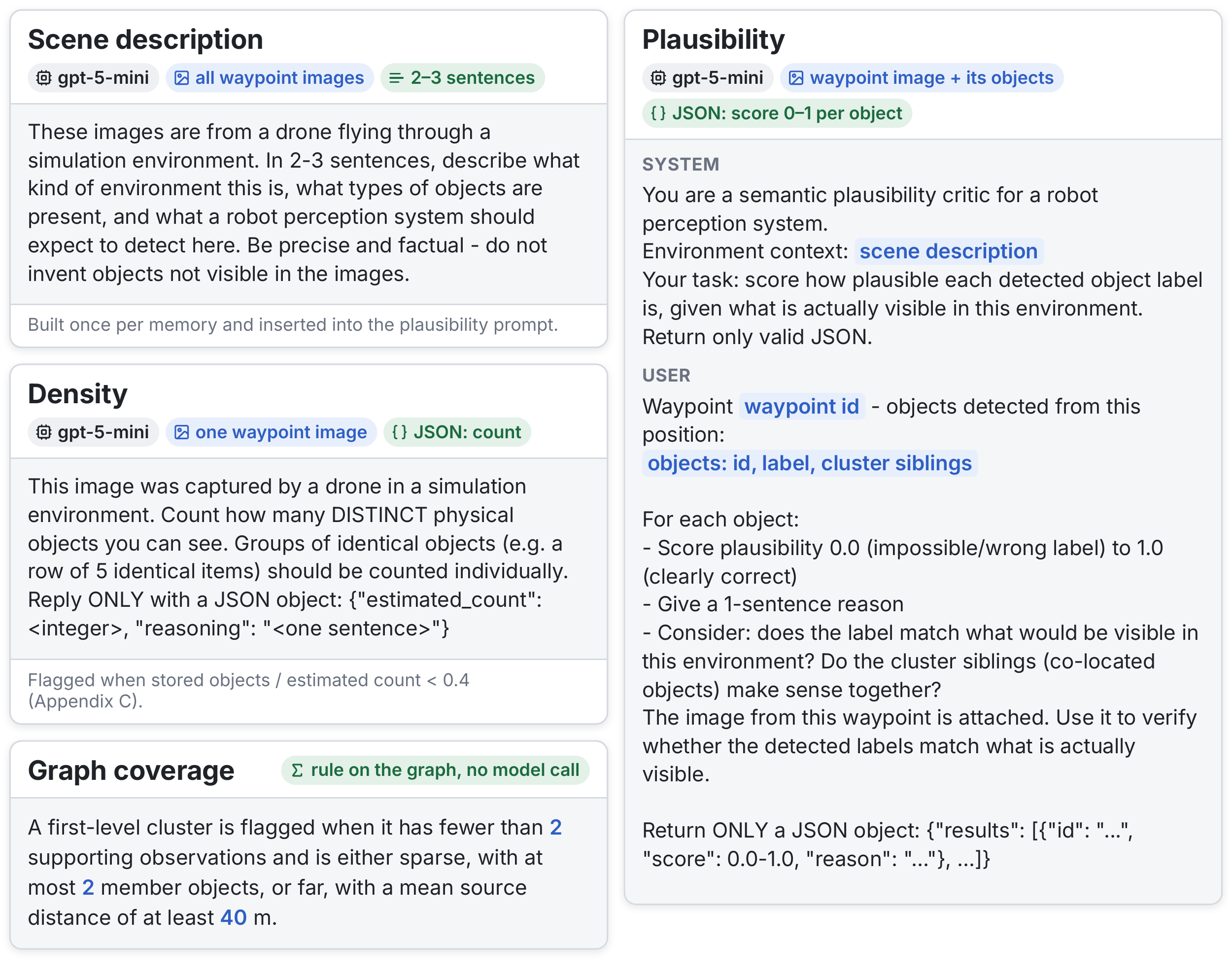}
\caption{Critic prompts. The scene description is generated once per memory and inserted into the plausibility prompt. Density and plausibility are separate calls per waypoint, while graph coverage makes no model call and is shown as the rule it applies.}
\label{fig:prompt-critic}
\end{figure}

\emph{Plausibility} has the VLM score every stored object on $[0,1]$ (prompts in Figure~\ref{fig:prompt-critic}) by whether its label is visually supported in context, and flags entries below $0.6$ as well as entries left unscored, so an object the critic could not assess is treated as unreliable. \emph{Density} estimates how many objects are visible in a waypoint image and flags the waypoint when the memory stores fewer than $0.4$ times that estimate. When the estimate is zero, the ratio is undefined: in the main results such a waypoint is marked unassessable, while in the critic ablation it is flagged whenever memory stores objects from it, because a frame that looks empty at a waypoint with stored objects points to a failed view. \emph{Graph coverage} flags a first-level cluster with fewer than two contributing observations when it is also sparse, with at most two member nodes, or far, with a mean source distance of at least $40$~m. The distance threshold matches the $40$~m depth limit, and requiring sparsity or distance in addition to low support keeps the rule from flagging every small cluster in a sparse scene.

The critic ablation reports recall without precision because precision is undefined after a degraded Episode~1, where every assessed cluster is poorly recovered. On Hangar, for example, clusters hold one to six objects against $17$ to $115$ oracle objects within $45$~m, and no cluster on any map holds even a third of the oracle objects near it. With no adequately recovered regions, the critic cannot produce a false positive, so precision would be $1.0$ by construction. Recall remains informative because it measures how many of the damaged regions each signal finds, and precision becomes measurable once correction restores adequately recovered regions.

\subsection{Question construction and scoring}

Each question asks about the region seen from one source waypoint and counts every stored object inside that region, whichever viewpoint observed it, so a loss at that waypoint shows up in its own questions instead of being averaged over the map. The region is a circle centered on the mean position of the oracle objects visible from the waypoint, with a radius equal to their largest offset from that center along either ground axis and at least $15$~m, which gives radii of $15$ to $27$~m. The ground truth and the memory answers count objects inside the same circle and only in the same shared categories, so neither side is credited for area or categories the other does not count. Repeated views of one object do not inflate memory counts, because detections of one category within $2$~m of each other from different observations count once, while detections from the same image are never combined, since one image does not show the same object twice.

\emph{Inventory} asks how many objects were in the region and how many of each of its up to three most common categories. Counts use seven ranges from \emph{none} through \emph{more than 20}, because coarser ranges would give memories with, for example, eight and fifty objects the same \emph{many~(10+)} answer. \emph{Dominant type} asks which category was most common, only when one category leads, since a tie has no single correct answer, and whether each of the most common categories was present, which captures views that keep incidental objects but lose their main content. \emph{Type richness} asks how many distinct categories were in the region, in four ranges (one, 2--3, 4--6, more than 6), since diversity can collapse even when some objects remain. \emph{Rare presence} asks whether a category that occurs elsewhere on the map but not in the region was there, and whether a category the oracle records exactly once in the region appears exactly once, since a single object is the first to disappear from a degraded view. Presence and absence questions are generated in pairs, so a memory that answers every existence question the same way cannot score well, and the absent categories are drawn at random with a fixed seed.

We keep the set to four families, which gives every waypoint eight to ten questions, because further families would mostly re-ask what these already measure, such as the count or presence of a single type, while thinning out the questions at each waypoint, which would let a few questions move the scores up or down.

\begin{figure}[t!]
\centering
\includegraphics[width=\linewidth]{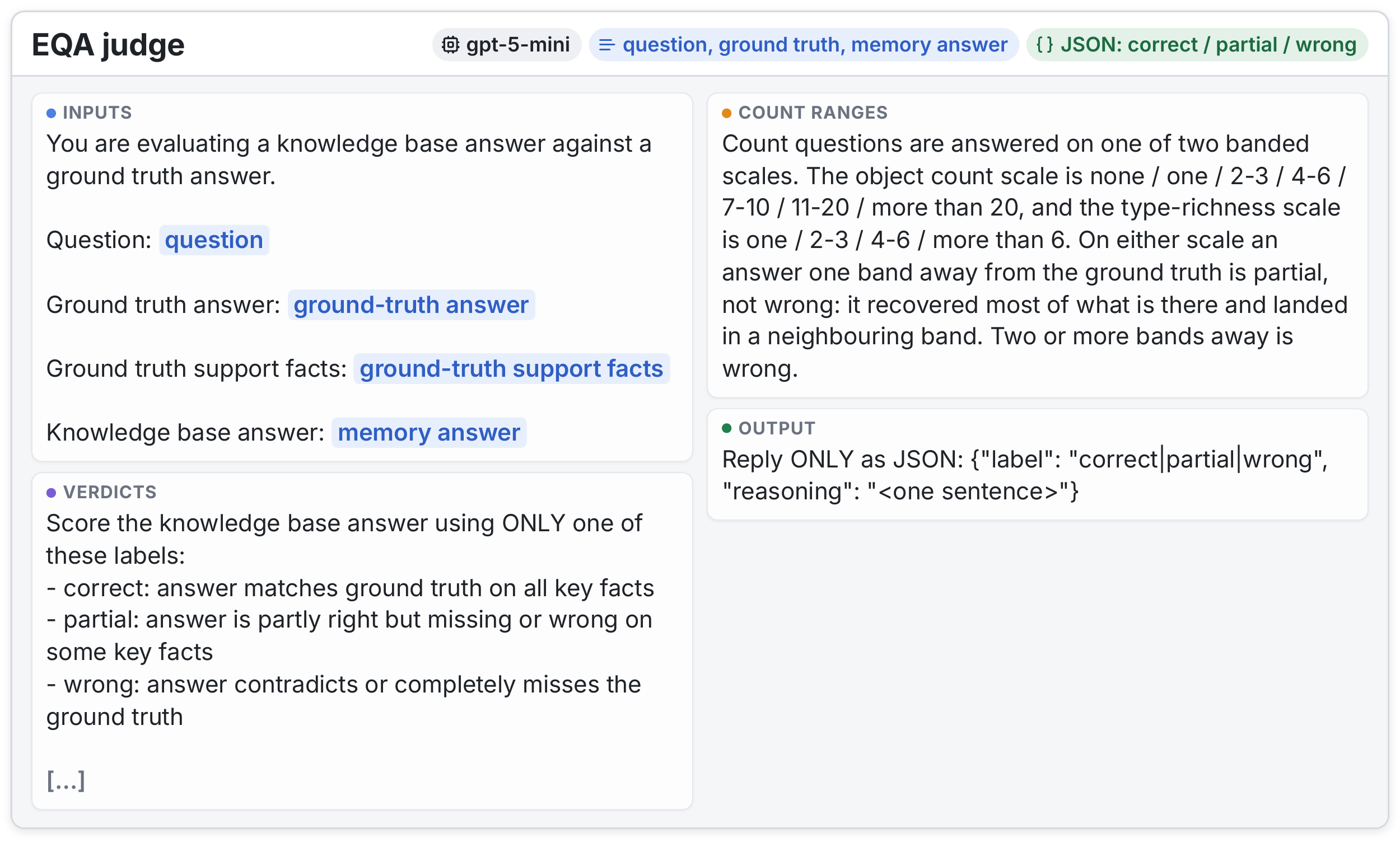}
\caption{EQA judge prompt. The same prompt scores answers from $H^{(1)}$ and $H^{(2)}$ against the oracle ground truth, and each condition is scored three times.}
\label{fig:prompt-judge}
\end{figure}

A language-model judge (Figure~\ref{fig:prompt-judge}) scores the answers from $H^{(1)}$ and $H^{(2)}$ on the same fixed question set, labeling each answer \emph{correct}~($1.0$), \emph{partial}~($0.5$), or \emph{wrong}~($0.0$), and a count answer one range from the reference receives partial credit. Questions that fail to score because of a judge or retrieval error are excluded from the mean and reported separately.

\begin{figure}[t!]
\centering
\includegraphics[width=\linewidth]{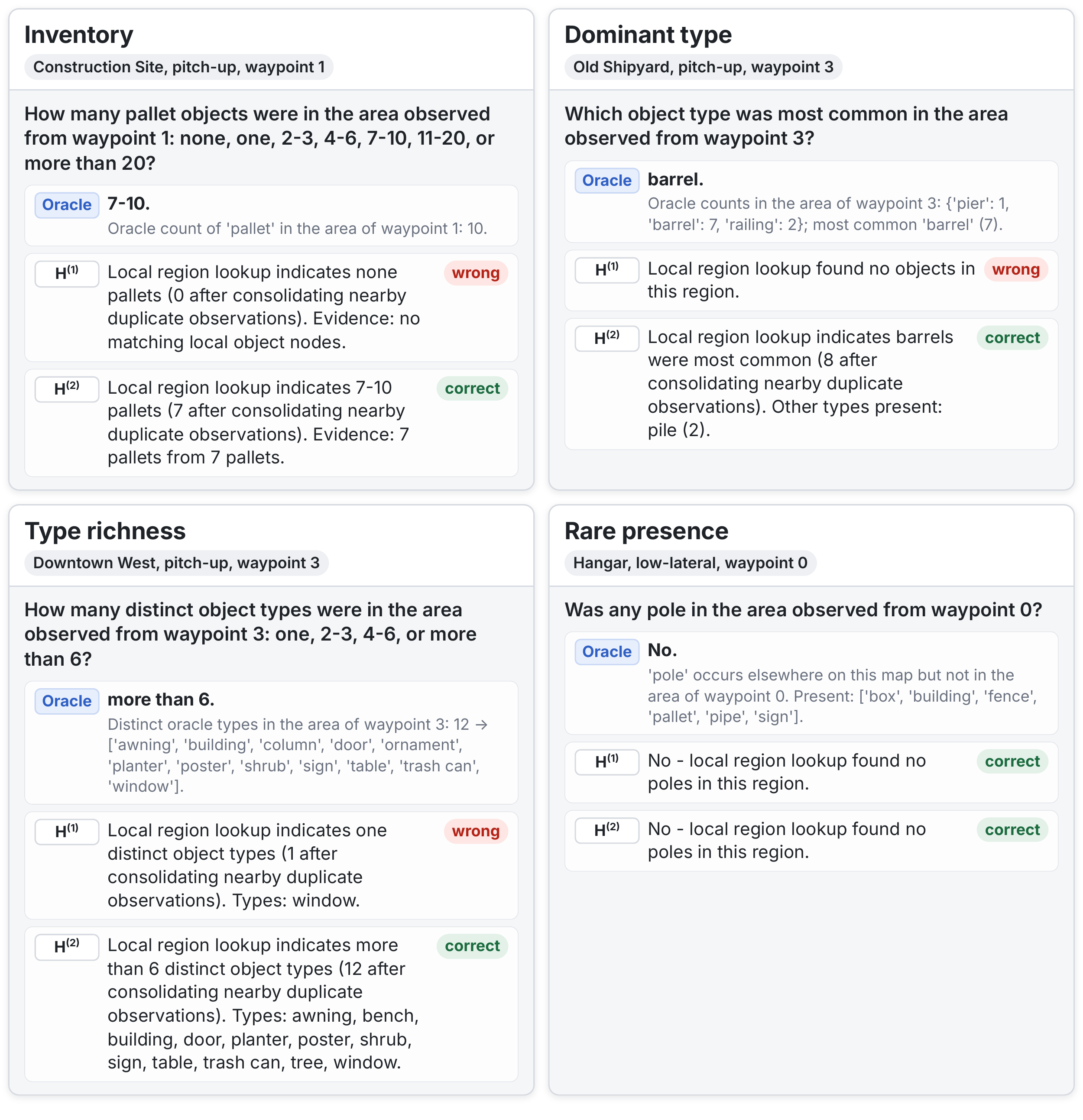}
\caption{One fixed-set question per family, with the oracle answer and the answers from $H^{(1)}$ and $H^{(2)}$. Memory answers are shown verbatim, and \textnormal{[\ldots]} marks a shortened evidence list. In the rare-presence example, both memories correctly report no pole at a waypoint where the map has poles elsewhere.}
\label{fig:eqa-examples}
\end{figure}

Figure~\ref{fig:eqa-examples} shows one question from each family before and after correction, chosen among questions that received the same verdicts in all three scoring passes, so the examples reflect the memories rather than variation in the judge.

\subsection{Question set composition}

The fixed set contains 158 questions: 40 on Old~Shipyard, 38 on Hangar, 40 on Construction~Site, and 40 on Downtown~West. Every waypoint receives one type-richness question and up to four inventory questions, its total count and the counts of up to three most common categories, so each map contributes four type-richness and 16 inventory questions, except Hangar, where one waypoint shows only two categories and receives 15. The rest are three or four dominant-type questions per map, as ties are skipped, seven or eight paired presence and absence questions each, and three exactly-once questions on Old~Shipyard.

Questions are not phrased with oracle object names alone, since these are simulator asset names that a memory built by the pipeline may never contain. We therefore restrict questions to categories present in both the oracle and the undegraded reference KB of each map, which leaves 8 categories on Old~Shipyard, 8 on Hangar, 15 on Construction~Site, and 17 on Downtown~West. The reference KB fixes only which categories questions may use, while the answers come from the oracle and the same categories apply to every memory scored on that map.

Each experimental condition is collected in a single flight and scored three times, and we report the mean. Across all conditions in the paper, the standard deviation across repeats averages $0.0028$, small relative to the reported changes in memory quality. As expected, perception itself also varies between flights: across six repeated Episode~1 collections from identical undegraded poses on one map, the oracle judge's memory-quality score stayed within $0.05$, with the differences coming from SAM2 masks, caption wording, and rendered exposure.

\subsection{Label agreement}
\label{app:label-agreement}

Clustering and the oracle judge both need to decide whether two labels name the same type of object, and both do so by comparing \texttt{text-embedding-3-small} embeddings by cosine similarity with a threshold of $0.40$. On the calibration set, this threshold accepted $16$ of $17$ same-object phrasing pairs and none of the $10$ different-object pairs. The highest different-object similarity was $0.357$, and the one same-object pair below the threshold scored $0.373$.

\section{Observation and Viewpoint Details}
\label{app:collection-details}

\subsection{Simulator and capture configuration}

The perception and oracle pipelines need different sensor channels, hence each uses its own AirSim capture profile, and the simulator is restarted when switching between them because capture settings are read only at startup. The perception profile exposes scene RGB and perspective depth, and the oracle profile adds planar depth and a per-actor segmentation channel for visibility testing. Both render at $1280\times720$ with a $90^\circ$ horizontal field of view, corresponding to a $58.7^\circ$ vertical field of view and a $29.4^\circ$ vertical half-angle, so the geometry and imagery of the two profiles can be compared directly. Hangar's source poses face shadowed areas under a bright sky, where automatic exposure leaves the ground nearly black, so on that map every perception frame of every episode, including the undegraded reference, is denoised and then brightened by a gamma and gain correction that applies only when the frame is dark.

\subsection{Motion and capture}

For each source or revisit pose, the vehicle climbs to a $20$~m transit altitude, translates to the target, descends to capture height, sets yaw on the airframe and pitch on the camera gimbal, stabilizes, and records one observation. Pitch goes through the gimbal because the flight controller re-levels the airframe within roughly two seconds, which makes body-pitch commands unsuitable for a stable capture pose. No observations are recorded in transit, so coverage depends on the evaluated viewpoints and not on the route between them.

\subsection{Viewpoint conditions}

The undegraded condition flies each hand-selected pose unchanged, while pitch-up places the bottom of the image $11.3^\circ$ above the horizon, which with the $29.4^\circ$ vertical half-angle gives a camera pitch of $+40.7^\circ$ on every map. Defining the degradation by the image boundary rather than by a fixed pitch offset keeps the loss of near-field coverage comparable across source poses that start at different pitches. Low-lateral lowers the camera by $3.3$--$3.7$~m on Hangar, Construction~Site, and Downtown~West, and by $8.8$--$9.1$~m per waypoint on Old~Shipyard, where uneven terrain means a single drop does not reach the same ground clearance, and turns the camera $45^\circ$ off its intended heading. We do not use a downward-pitch degradation, because at these capture heights it mostly replaces the scene with nearby pavement that still segments cleanly, while pitch-up removes the objects in the lower part of the view. Because the critic and planner never see the degradation, correction fails where they misplace the losses or where the revisits and merge do not turn new views into correct answers. Hangar under low-lateral gains only $0.017$, its untargeted revisits lose $0.022$, and the combined critic misses $12$ of the $60$ poorly recovered regions in its ablation.

\subsection{Transit on Downtown~West}

Tree geometry on Downtown~West makes the standard flight controller unreliable over the long routes between revisit poses, hence, revisit episodes on that map travel at transit altitude through a sequence of directly set poses. Descents to capture height still use the flight controller, as in Episode~1, so the vehicle holds its capture height the same way in every episode, and the transit mode changes only how the vehicle moves between capture poses, not how observations are recorded.

\subsection{Map and waypoint selection}

The four maps span different levels of clutter and enclosure and still stay small enough to inspect individual correction traces, since every map-condition evaluation runs the whole pipeline. Within each map, the four source waypoints were selected for object diversity, object density, and regional isolation, to give each one a distinct local area that makes a meaningful revisit target.

\subsection{Oracle collection}

The oracle matches the perception pipeline in image resolution, field of view, source locations, and $40$~m depth limit, but always observes from the undegraded camera pose. To keep the oracle from recording objects hidden behind walls, visibility is tested by ray casting against the segmentation channel. Scene queries still return elements that are not objects, such as lights, cameras, terrain, and decals, which are removed by name filtering (Sec.~\ref{sec:oracle}). The remaining asset names differ from ordinary object names, which is why they pass through the same open-vocabulary normalization as the memory's names (Appendix~\ref{app:perception}).

\end{document}

%% file: math_commands.tex
\usepackage{amsmath,amsfonts,bm}

\def\eqref#1{equation~\ref{#1}}
\def\1{\bm{1}}

\DeclareMathAlphabet{\mathsfit}{\encodingdefault}{\sfdefault}{m}{sl}
\SetMathAlphabet{\mathsfit}{bold}{\encodingdefault}{\sfdefault}{bx}{n}

%% file: iclr2027_conference.bib
@misc{embodiedrag2024,
  title         = {{Embodied-RAG: General Non-parametric Embodied Memory for Retrieval and Generation}},
  author        = {Xie, Quanting and Min, So Yeon and Ji, Pengliang and Yang, Yue and Zhang, Tianyi and Xu, Kedi and Bajaj, Aarav and Salakhutdinov, Ruslan and Johnson-Roberson, Matthew and Bisk, Yonatan},
  year          = {2024},
  howpublished  = {arXiv preprint arXiv:2409.18313},
  eprint        = {2409.18313},
  archivePrefix = {arXiv} }

@inproceedings{lewis2020rag,
  title         = {{Retrieval-Augmented Generation for Knowledge-Intensive NLP Tasks}},
  author        = {Lewis, Patrick and Perez, Ethan and Piktus, Aleksandra and Petroni, Fabio and Karpukhin, Vladimir and Goyal, Naman and K{\"u}ttler, Heinrich and Lewis, Mike and Yih, Wen-tau and Rockt{\"a}schel, Tim and Riedel, Sebastian and Kiela, Douwe},
  booktitle     = {Advances in Neural Information Processing Systems},
  volume        = {33},
  pages         = {9459--9474},
  publisher     = {Curran Associates, Inc.},
  year          = {2020},
  eprint        = {2005.11401},
  archivePrefix = {arXiv} }

@misc{graphrag2024,
  title         = {{From Local to Global: A Graph RAG Approach to Query-Focused Summarization}},
  author        = {Edge, Darren and Trinh, Ha and Cheng, Newman and Bradley, Joshua and Chao, Alex and Mody, Apurva and Truitt, Steven and Metropolitansky, Dasha and Ness, Robert Osazuwa and Larson, Jonathan},
  year          = {2024},
  howpublished  = {arXiv preprint arXiv:2404.16130},
  eprint        = {2404.16130},
  archivePrefix = {arXiv} }

@inproceedings{mindpalace2025,
  title         = {{Enter the Mind Palace: Reasoning and Planning for Long-term Active Embodied Question Answering}},
  author        = {Ginting, Muhammad Fadhil and Kim, Dong-Ki and Meng, Xiangyun and Reinke, Andrzej Marek and Krishna, Bandi Jai and Kayhani, Navid and Peltzer, Oriana and Fan, David and Shaban, Amirreza and Kim, Sung-Kyun and Kochenderfer, Mykel and Agha-mohammadi, Ali-akbar and Omidshafiei, Shayegan},
  booktitle     = {Proceedings of The 9th Conference on Robot Learning},
  series        = {Proceedings of Machine Learning Research},
  volume        = {305},
  pages         = {5072--5106},
  publisher     = {PMLR},
  year          = {2025},
  eprint        = {2507.12846},
  archivePrefix = {arXiv} }

@inproceedings{remembr2024,
  title         = {{ReMEmbR: Building and Reasoning Over Long-Horizon Spatio-Temporal Memory for Robot Navigation}},
  author        = {Anwar, Abrar and Welsh, John and Biswas, Joydeep and Pouya, Soha and Chang, Yan},
  booktitle     = {2025 IEEE International Conference on Robotics and Automation (ICRA)},
  pages         = {2838--2845},
  publisher     = {IEEE},
  year          = {2025},
  doi           = {10.1109/ICRA55743.2025.11127706},
  eprint        = {2409.13682},
  archivePrefix = {arXiv} }

@misc{crag2024,
  title         = {{Corrective Retrieval Augmented Generation}},
  author        = {Yan, Shi-Qi and Gu, Jia-Chen and Zhu, Yun and Ling, Zhen-Hua},
  year          = {2024},
  howpublished  = {arXiv preprint arXiv:2401.15884},
  eprint        = {2401.15884},
  archivePrefix = {arXiv} }

@inproceedings{selfrag2024,
  title         = {{Self-RAG: Learning to Retrieve, Generate, and Critique through Self-Reflection}},
  author        = {Asai, Akari and Wu, Zeqiu and Wang, Yizhong and Sil, Avirup and Hajishirzi, Hannaneh},
  booktitle     = {The Twelfth International Conference on Learning Representations},
  year          = {2024},
  eprint        = {2310.11511},
  archivePrefix = {arXiv} }

@inproceedings{reflect2023,
  title         = {{REFLECT: Summarizing Robot Experiences for Failure Explanation and Correction}},
  author        = {Liu, Zeyi and Bahety, Arpit and Song, Shuran},
  booktitle     = {Proceedings of The 7th Conference on Robot Learning},
  series        = {Proceedings of Machine Learning Research},
  volume        = {229},
  pages         = {3468--3484},
  publisher     = {PMLR},
  year          = {2023},
  eprint        = {2306.15724},
  archivePrefix = {arXiv} }

@misc{rag3dsg2026,
  title         = {{RAG-3DSG: Enhancing 3D Scene Graphs with Re-Shot Guided Retrieval-Augmented Generation}},
  author        = {Chang, Yue and Chen, Rufeng and Zhang, Zhaofan and Chen, Yi and Tian, Yifan and Xie, Sihong},
  year          = {2026},
  howpublished  = {arXiv preprint arXiv:2601.10168},
  note          = {Accepted to ECCV 2026},
  eprint        = {2601.10168},
  archivePrefix = {arXiv} }

@misc{graphpad2025,
  title         = {{GraphPad: Inference-Time 3D Scene Graph Updates for Embodied Question Answering}},
  author        = {Ali, Muhammad Qasim and Nair, Saeejith and Wong, Alexander and Cui, Yuchen and Chen, Yuhao},
  year          = {2025},
  howpublished  = {arXiv preprint arXiv:2506.01174},
  note          = {Presented at the CVPR 2025 Workshop on 3D-LLM/VLA: Bridging Language, Vision and Action in 3D Environments},
  eprint        = {2506.01174},
  archivePrefix = {arXiv} }

@misc{ogscene3d2026,
  title         = {{OGScene3D: Incremental Open-Vocabulary 3D Gaussian Scene Graph Mapping for Scene Understanding}},
  author        = {Zhu, Siting and Lu, Ziyun and Wang, Guangming and Huang, Chenguang and Chen, Yongbo and Chen, I-Ming and Burgard, Wolfram and Wang, Hesheng},
  year          = {2026},
  howpublished  = {arXiv preprint arXiv:2603.16301},
  eprint        = {2603.16301},
  archivePrefix = {arXiv} }

@misc{gsmem2026,
  title         = {{GSMem: 3D Gaussian Splatting as Persistent Spatial Memory for Zero-Shot Embodied Exploration and Reasoning}},
  author        = {Lu, Yiren and Du, Yi and Liu, Disheng and Zhou, Yunlai and Wang, Chen and Yin, Yu},
  year          = {2026},
  howpublished  = {arXiv preprint arXiv:2603.19137},
  eprint        = {2603.19137},
  archivePrefix = {arXiv} }

@inproceedings{airsim2017,
  title         = {{AirSim: High-Fidelity Visual and Physical Simulation for Autonomous Vehicles}},
  author        = {Shah, Shital and Dey, Debadeepta and Lovett, Chris and Kapoor, Ashish},
  booktitle     = {Field and Service Robotics},
  editor        = {Hutter, Marco and Siegwart, Roland},
  series        = {Springer Proceedings in Advanced Robotics},
  volume        = {5},
  pages         = {621--635},
  publisher     = {Springer International Publishing},
  year          = {2018},
  doi           = {10.1007/978-3-319-67361-5_40},
  eprint        = {1705.05065},
  archivePrefix = {arXiv} }

@inproceedings{sam2_2024,
  title         = {{SAM 2: Segment Anything in Images and Videos}},
  author        = {Ravi, Nikhila and Gabeur, Valentin and Hu, Yuan-Ting and Hu, Ronghang and Ryali, Chaitanya and Ma, Tengyu and Khedr, Haitham and R{\"a}dle, Roman and Rolland, Chloe and Gustafson, Laura and Mintun, Eric and Pan, Junting and Alwala, Kalyan Vasudev and Carion, Nicolas and Wu, Chao-Yuan and Girshick, Ross and Doll{\'a}r, Piotr and Feichtenhofer, Christoph},
  booktitle     = {The Thirteenth International Conference on Learning Representations},
  year          = {2025},
  eprint        = {2408.00714},
  archivePrefix = {arXiv} }

@article{khoshelham2012,
  title         = {{Accuracy and Resolution of Kinect Depth Data for Indoor Mapping Applications}},
  author        = {Khoshelham, Kourosh and Oude Elberink, Sander},
  journal       = {Sensors},
  volume        = {12},
  number        = {2},
  pages         = {1437--1454},
  year          = {2012},
  doi           = {10.3390/s120201437} }

@inproceedings{nguyen2012,
  title         = {{Modeling Kinect Sensor Noise for Improved 3D Reconstruction and Tracking}},
  author        = {Nguyen, Chuong V. and Izadi, Shahram and Lovell, David},
  booktitle     = {2012 Second International Conference on 3D Imaging, Modeling, Processing, Visualization \& Transmission},
  pages         = {524--530},
  publisher     = {IEEE},
  year          = {2012},
  doi           = {10.1109/3DIMPVT.2012.84} }

@inproceedings{henriques2018mapnet,
  title         = {{MapNet: An Allocentric Spatial Memory for Mapping Environments}},
  author        = {Henriques, Jo{\~a}o F. and Vedaldi, Andrea},
  booktitle     = {2018 IEEE/CVF Conference on Computer Vision and Pattern Recognition},
  pages         = {8476--8484},
  publisher     = {IEEE},
  year          = {2018},
  doi           = {10.1109/CVPR.2018.00884} }

@inproceedings{ahn2022can,
  title         = {{Do As I Can, Not As I Say: Grounding Language in Robotic Affordances}},
  author        = {Ahn, Michael and Brohan, Anthony and Brown, Noah and Chebotar, Yevgen and Cortes, Omar and David, Byron and Finn, Chelsea and Fu, Chuyuan and Gopalakrishnan, Keerthana and Hausman, Karol and Herzog, Alex and Ho, Daniel and Hsu, Jasmine and Ibarz, Julian and Ichter, Brian and Irpan, Alex and Jang, Eric and Jauregui Ruano, Rosario and Jeffrey, Kyle and Jesmonth, Sally and Joshi, Nikhil J. and Julian, Ryan and Kalashnikov, Dmitry and Kuang, Yuheng and Lee, Kuang-Huei and Levine, Sergey and Lu, Yao and Luu, Linda and Parada, Carolina and Pastor, Peter and Quiambao, Jornell and Rao, Kanishka and Rettinghouse, Jarek and Reyes, Diego and Sermanet, Pierre and Sievers, Nicolas and Tan, Clayton and Toshev, Alexander and Vanhoucke, Vincent and Xia, Fei and Xiao, Ted and Xu, Peng and Xu, Sichun and Yan, Mengyuan and Zeng, Andy},
  booktitle     = {Proceedings of The 6th Conference on Robot Learning},
  series        = {Proceedings of Machine Learning Research},
  volume        = {205},
  pages         = {287--318},
  publisher     = {PMLR},
  year          = {2023},
  eprint        = {2204.01691},
  archivePrefix = {arXiv} }

@inproceedings{driess2023palm,
  title         = {{PaLM-E: An Embodied Multimodal Language Model}},
  author        = {Driess, Danny and Xia, Fei and Sajjadi, Mehdi S. M. and Lynch, Corey and Chowdhery, Aakanksha and Ichter, Brian and Wahid, Ayzaan and Tompson, Jonathan and Vuong, Quan and Yu, Tianhe and Huang, Wenlong and Chebotar, Yevgen and Sermanet, Pierre and Duckworth, Daniel and Levine, Sergey and Vanhoucke, Vincent and Hausman, Karol and Toussaint, Marc and Greff, Klaus and Zeng, Andy and Mordatch, Igor and Florence, Pete},
  booktitle     = {Proceedings of the 40th International Conference on Machine Learning},
  series        = {Proceedings of Machine Learning Research},
  volume        = {202},
  pages         = {8469--8488},
  publisher     = {PMLR},
  year          = {2023},
  eprint        = {2303.03378},
  archivePrefix = {arXiv} }

@inproceedings{huang2022inner,
  title         = {{Inner Monologue: Embodied Reasoning through Planning with Language Models}},
  author        = {Huang, Wenlong and Xia, Fei and Xiao, Ted and Chan, Harris and Liang, Jacky and Florence, Pete and Zeng, Andy and Tompson, Jonathan and Mordatch, Igor and Chebotar, Yevgen and Sermanet, Pierre and Jackson, Tomas and Brown, Noah and Luu, Linda and Levine, Sergey and Hausman, Karol and Ichter, Brian},
  booktitle     = {Proceedings of The 6th Conference on Robot Learning},
  series        = {Proceedings of Machine Learning Research},
  volume        = {205},
  pages         = {1769--1782},
  publisher     = {PMLR},
  year          = {2023},
  eprint        = {2207.05608},
  archivePrefix = {arXiv} }

@inproceedings{gu2024conceptgraphs,
  title         = {{ConceptGraphs: Open-Vocabulary 3D Scene Graphs for Perception and Planning}},
  author        = {Gu, Qiao and Kuwajerwala, Ali and Morin, Sacha and Jatavallabhula, Krishna Murthy and Sen, Bipasha and Agarwal, Aditya and Rivera, Corban and Paul, William and Ellis, Kirsty and Chellappa, Rama and Gan, Chuang and de Melo, Celso Miguel and Tenenbaum, Joshua B. and Torralba, Antonio and Shkurti, Florian and Paull, Liam},
  booktitle     = {2024 IEEE International Conference on Robotics and Automation (ICRA)},
  pages         = {5021--5028},
  publisher     = {IEEE},
  year          = {2024},
  doi           = {10.1109/ICRA57147.2024.10610243} }

@misc{singh2025gpt5,
  title        = {{OpenAI GPT-5 System Card}},
  author       = {{OpenAI}},
  year         = {2025},
  eprint       = {2601.03267},
  archivePrefix = {arXiv},
  primaryClass = {cs.CL},
  doi          = {10.48550/arXiv.2601.03267}
}

@inproceedings{dengler2021online,
  title     = {Online Object-Oriented Semantic Mapping and Map Updating},
  author    = {Dengler, Nils and Zaenker, Tobias and Verdoja, Francesco and Bennewitz, Maren},
  booktitle = {2021 European Conference on Mobile Robots (ECMR)},
  pages     = {1--7},
  publisher = {IEEE},
  year      = {2021},
  doi       = {10.1109/ECMR50962.2021.9568817},
  eprint    = {2011.06895},
  archivePrefix = {arXiv}
}

@article{matez2022ltc,
  title   = {{LTC-Mapping, Enhancing Long-Term Consistency of Object-Oriented Semantic Maps in Robotics}},
  author  = {Matez-Bandera, Jose-Luis and Fernandez-Chaves, David and Ruiz-Sarmiento, Jose-Raul and Monroy, Javier and Petkov, Nicolai and Gonzalez-Jimenez, Javier},
  journal = {Sensors},
  volume  = {22},
  number  = {14},
  pages   = {5308},
  year    = {2022},
  doi     = {10.3390/s22145308}
}

@misc{openai2024embeddings,
  title        = {New embedding models and {API} updates},
  author       = {{OpenAI}},
  year         = {2024},
  month        = jan,
  howpublished = {\url{https://openai.com/index/new-embedding-models-and-api-updates/}}
}

@misc{airsimdocs,
  title        = {{AirSim} Documentation: {APIs} and Image {APIs}},
  author       = {{Microsoft}},
  year         = {2026},
  note         = {Accessed October 2026},
  howpublished = {\url{https://microsoft.github.io/AirSim/apis/}}
}
